\documentclass[letterpaper]{article} 
\usepackage{aaai2026}  
\usepackage{times}  
\usepackage{helvet}  
\usepackage{courier}  
\usepackage[hyphens]{url}  
\usepackage{graphicx} 
\usepackage{natbib}  
\usepackage{caption} 
\usepackage{algorithm}
\usepackage{algorithmic}
\usepackage{multirow}
\usepackage{makecell}
\usepackage{booktabs}
\usepackage{amsmath,amssymb,amsfonts}
\usepackage{subfig}

\usepackage{newfloat}
\usepackage{listings}
\DeclareCaptionStyle{ruled}{labelfont=normalfont,labelsep=colon,strut=off} 
\floatstyle{ruled}
\newfloat{listing}{tb}{lst}{}
\floatname{listing}{Listing}
\title{RAC-DMVC: Reliability-Aware Contrastive Deep Multi-View Clustering under Multi-Source Noise}
\author {
    Shihao Dong\textsuperscript{\rm 1},
    Yue Liu\textsuperscript{\rm 2},
    Xiaotong Zhou\textsuperscript{\rm 1},
    Yuhui Zheng\textsuperscript{\rm 3}\thanks{Corresponding author},
    Huiying Xu\textsuperscript{\rm 4}, 
    Xinzhong Zhu\textsuperscript{\rm 4}
}
\affiliations {
    \textsuperscript{\rm 1}School of Computer Science, Nanjing University of Information Science and Technology, Nanjing, China.\\
    \textsuperscript{\rm 2}School of Computing, National University of Singapore, Singapore.\\
    \textsuperscript{\rm 3}Key Laboratory of Tibetan Information Processing, Ministry of Education, Qinghai Normal University. Xining, China.\\
    \textsuperscript{\rm 4}School of Computer Science and Technology, Zhejiang Normal University. Jinhua, China.\\
    \{dongshihao, xiaotong\_zhou\}@nuist.edu.cn, yliu@u.nus.edu, zhengyh@vip.126.com, \{xhy, zxz\}@zjnu.edu.cn
}

\usepackage{bibentry}

\begin{document}

\maketitle

\begin{abstract}
Multi-view clustering (MVC), which aims to separate the multi-view data into distinct clusters in an unsupervised manner, is a fundamental yet challenging task. To enhance its applicability in real-world scenarios, this paper addresses a more challenging task: MVC under multi-source noises, including missing noise and observation noise. To this end, we propose a novel framework, Reliability-Aware Contrastive Deep Multi-View Clustering (RAC-DMVC), which constructs a reliability graph to guide robust representation learning under noisy environments. Specifically, to address observation noise, we introduce a cross-view reconstruction to enhances robustness at the data level, and a reliability-aware noise contrastive learning to mitigates bias in positive and negative pairs selection caused by noisy representations. To handle missing noise, we design a dual-attention imputation to capture shared information across views while preserving view-specific features. In addition, a self-supervised cluster distillation module further refines the learned representations and improves the clustering performance. Extensive experiments on five benchmark datasets demonstrate that RAC-DMVC outperforms SOTA methods on multiple evaluation metrics and maintains excellent performance under varying ratios of noise.
\end{abstract}

\begin{links}
    \link{Code}{https://github.com/LouisDong95/RAC-DMVC}
\end{links}

\section{Introduction}
In recent years, advances in sensing and data acquisition technologies have facilitated the widespread use of multi-modal and multi-view data across a variety of practical applications. Multi-view data captures complementary information from heterogeneous sources (e.g., images, text, signals) and offers multiple perspectives on the same entity, thereby enhancing data expressiveness. Effectively integrating such information has become critical in domains such as industrial inspection~\cite{0020PZYWW23,AsadAMJAYL25}, social network analysis~\cite{chen2024multi,HuangZLLHZ18}, healthcare~\cite{li2025curegraph,holm2025amvae}, and biomedicine~\cite{cui2025towards,rao2025multimodal}. Consequently, clustering multi-view data to uncover its latent structure and patterns has attracted increasing attention from the research community.

As a key technique for analyzing multi-view data, MVC aims to leverage complementary information from diverse perspectives or modalities to perform unsupervised clustering. Early MVC methods were primarily rooted in traditional machine learning approaches, including subspace learning~\cite{gao2015multi, cao2015diversity, kang2020large}, non-negative matrix factorization~\cite{liu2013multi, zhao2017multi, huang2020auto}, and graph-based learning~\cite{tang2020cgd, wang2019gmc}. These methods promote the effective fusion of multi-view information by designing consistency constraints or shared latent representations. With the advent of deep learning, deep multi-view clustering (DMVC)~\cite{li2019deep,0001PCZCP024,zhang2025incomplete,zhang2025multi} has emerged, offering superior performance under complex data distributions through powerful end-to-end feature learning. Although multi-view clustering methods have achieved remarkable progress, they still encounter significant challenges caused by noise in real-world applications.

In the acquisition and processing of multi-view data, the presence of noise is inevitable—particularly view missing, which often results from acquisition errors, sensor failures, or communication interruptions. Existing methods for handling missing views include imputation-based methods~\cite{TangL22,JinWDLZ23,Li0Y0P023,PuCC0PHY024,YuanSZWYYR25,ChaoJC24} which aim to restore the missing views using available information, and non-imputation methods~\cite{LinGLBLP23,XuL0PMS022,Lu0YPH024,FengSWGTD24}, which avoid the additional noise caused by imputation through cross-view prediction. Although existing methods perform well in handling missing noise, they often overlook observation noise, such as lighting artifacts in nighttime images, background interference in audio signals, or blurred handwriting in text data. In real-world scenarios, observation noise is not only more prevalent than missing noise but also more likely to be overlooked. Noisy views often fail to provide useful information and may even degrade the quality of fused features, posing a significant challenge to multi-view clustering. Therefore, developing a unified and robust framework to jointly handle both missing and observation noise has become a crucial step toward enhancing the reliability and applicability of MVC in practical noisy environments.

To effectively address the challenge of multi-source noise in MVC, we propose a Reliability-Aware Contrastive Deep Multi-View Clustering framework. This framework constructs a reliability graph based on inter-sample similarity to guide noise contrastive learning and dual-attention imputation to mitigate the adverse effects of observation noise and missing noise on clustering performance. Specifically, we first cross-view reconstruct clean view from noisy views, enhancing robustness at the data level. Then, a feature similarity graph is constructed to simultaneously guide the contrastive learning against observation noise and the imputation process for missing data. For observation noise, the graph serves as the weights of positive and negative pairs for reliability-aware noise contrastive to mitigate the bias in positive and negative pair selection caused by noise. For missing noise, we design a dual-attention imputation to jointly model view-specific features and cross-view shared representations to assign adaptive attention weights, thereby improving imputation accuracy. Finally, we incorporate a self-supervised cluster distillation strategy to leverage the cluster distribution information to regularize and guide the learning of view representation learning. This collaborative design significantly improves the clustering performance under noisy scenarios. Our main contributions are as follows:
\begin{itemize}
\item 
{For the first time, we systematically consider the impact of missing and observation noise in the MVC task and propose a reliability-aware contrastive deep multi-view clustering method.}
\item 
{By constructing a reliable graph to guide noise contrastive learning and dual-attention imputation, our method effectively mitigates the effects of both observation and missing noise.}
\item 
{On five widely used multi-view datasets, the performance of our method under various noise ratios is systematically evaluated, and the results fully verify the superiority of our method compared with the SOTA methods.}
\end{itemize}

\section{Related Works}
\subsection{Deep Multi-view Clustering}
In recent years, DMVC has been greatly developed due to the powerful feature extraction ability of deep nerual networks. According to the clustering method can be divided into three categories: 1) DEC-based: These approaches are typically built upon Deep Embedded Clustering, which jointly optimizes feature learning and clustering in a unified framework. DAMC~\cite{li2019deep} proposes a deep adversarial MVC network to learn the intrinsic structure embedding in multi-view data. AIMC~\cite{xu2019adversarial} extends it to handle incomplete multi-view data. Multi-VAE~\cite{xu2021multi} proposes a VAE-based MVC framework by learning disentangled visual representations. 2) Subspace clustering-based: These methods aim to project multi-view data into a common latent subspace where self-expressiveness holds. DMSC~\cite{abavisani2018deep} inserts self-expression layers in AEs and compares the clustering performance of the fusion methods at different stages. DMVSSC~\cite{TangTWFW18} proposes a deep multi-view sparse subspace clustering model consisting of convolutional-AE and self-expressive module. 3) GNN-based: These methods utilize graph neural networks to model and aggregate multi-view relational information through graph structures. MAGCN~\cite{cheng2021multi} proposes a multi-view attribute graph convolution network model to handle node attributes and graph reconstruction. DFMVC~\cite{0001PCZCP024} proposes a dynamic graph fusion method to guide the learning of each view feature through the consistent features fused.

\subsection{Noisy Multi-view Learning}
Multi-view learning faces the challenge of multi-source noise in real-world scenarios, mainly including missing noise and non-missing noise. For missing noise, existing research is relatively mature, including: 
1) Imputation methods~\cite{TangL22,Yang00B0023,JinWDLZ23} use observable neighboring samples to achieve imputation, thereby retaining isomorphic structural information; Some works~\cite{Li0Y0P023,PuCC0PHY024,YuanSZWYYR25} perform imputation by calculating reliable prototypes to enhance the discriminability of imputation results. 2) Prediction methods~\cite{XuL0PMS022,LinGLBLP23,Lu0YPH024} adopt cross-view prediction strategies to avoid additional noise introduced by the interpolation process. In addition, DMVG~\cite{0001DWC0F024} introduces a conditional diffusion model to use available views as conditions to guide the generation of missing views.
For non-missing noise, it mainly includes observation noise~\cite{Xu0W0Z0024,yang2025automatically}, pseudo-label noise~\cite{SunLRDP025} and misalignment noise~\cite{Yang00B0023, GuoY00024}:
For observation noise, MVCAN~\cite{Xu0W0Z0024} combines feature representation with soft labels to optimize robust targets to improve noise resistance; AIRMVC~\cite{yang2025automatically} models noise recognition as anomaly detection tasks, introduces gaussian mixture models for anomaly modeling, and weakens the impact of noise through a hybrid rectification strategy, while combining contrastive learning to obtain a more robust feature representation.

\section{Methods}
\begin{figure*}[htbp]
    \centering
    \includegraphics[width=1\linewidth]{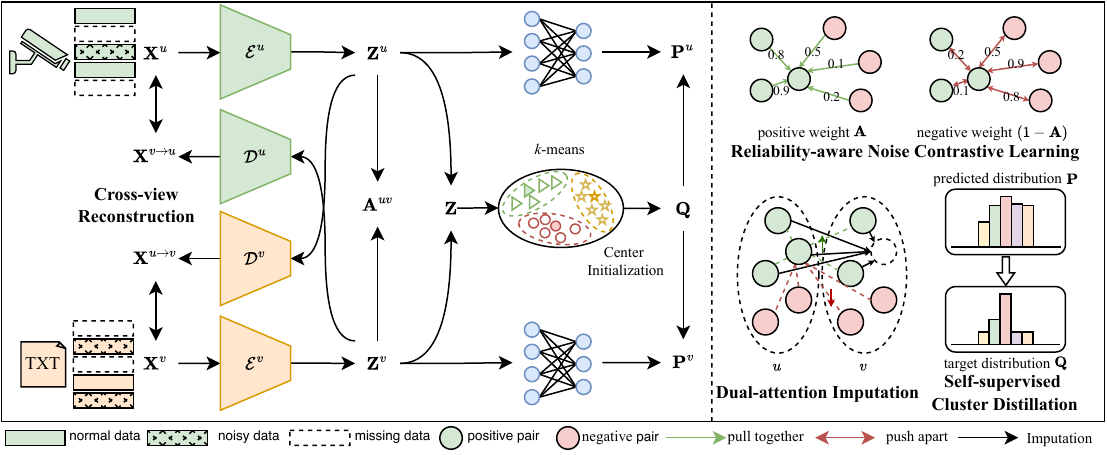}
    \caption{\textbf{Overview of the our pipeline.} The processing flow begins with cross-view reconstruction for robust representation. Subsequently, a similarity-based reliability graph is constructed to guide both contrastive learning and a dual-attention imputation module. Finally, self-supervised clustering distillation is applied to refine the view-specific representations.}
    \label{fig_pipline}
\end{figure*}

\subsection{Preliminaries}
Given a multi-view dataset $\mathcal{X}=\{\mathbf{X}^1, ..., \mathbf{X}^V\}$ of $N$ samples $V$ views,  $\mathbf{X}^v=\{\mathbf{x}^v_1, ..., \mathbf{x}^v_N\}\in \mathbb{R}^{d_v\times N}$, where $\mathbf{x}^v_i$ represents the $i$-th sample of the $v$-th view, and $d_v$ represents the dimension of the $v$-th view data. Due to the presence of noise, the multi-view data is divided into normal data $\mathbf{X}_{\mathrm{normal}}$ and noise data $\mathbf{X}_{\mathrm{noisy}}$, $\mathbf{X} = \mathbf{X}_{\mathrm{normal}}\bigcup\mathbf{X}_{\mathrm{noisy}}$. Due to the existence of multi-source noise, the noise includes missing noise $\overline{\mathbf{X}}$ and observation noise $\tilde{\mathbf{X}}$, $\mathbf{X}_{\mathrm{noisy}} = \overline{\mathbf{X}}\bigcup\tilde{\mathbf{X}}$, Let $\eta_m, \eta_n \in [0, 1]$ denote the ratios of missing noise and observation noise, respectively. Our objective is to divide the unlabeled data into $K$ clusters when there is complex noise in the multi-view data, so that the samples in the same cluster are closer and the samples in different clusters are farther away. The pipeline of our method is shown in the Fig.~\ref{fig_pipline}.

\subsection{Cross-view Reconstruction for Denoising}
In multi-view learning, heterogeneous data typically contain both view-specific features and shared correlation information across views. Existing methods often employ intra-view reconstruction via autoencoders to capture view-specific representations. However, this strategy suffers from two key limitations in noisy scenarios. 1) Overfitting to local noise: Noisy observations within a view are directly reconstructed, leading to corrupted latent features. 2) Lack of semantic integration: Independent view-wise encoding overlooks the cross-view complementarity and global consistency. To address these issues, we propose a cross-view reconstruction to guide the encoded features of one view to reconstruct the data from another view:
\begin{equation}
\ell_{\mathrm{crec}}(u, v) = 
\left\| \mathbf{X}^v - \mathcal{D}^v\left( \mathcal{E}^u(\tilde{\mathbf{X}}^u) \right) \right\|_2^2,
\end{equation}
\begin{equation}
\ell_{\mathrm{crec}} = \sum_{\substack{u,v \\ v \ne u}}^V\ell_{\mathrm{crec}}(u,v).
\end{equation}
Among them, $\mathcal{E}^u, \mathcal{D}^v$ represent the $u$-th encoder and the $v$-th decoder respectively. Through the cross-view reconstruction mechanism, not only can the observation noise be effectively fitted, but also the view-specific features can be retained while incorporating information between different views.

\subsection{Reliability-aware Noise Contrastive Learning}
Traditional multi-view contrastive learning methods usually regard the representations of the same sample under different views as positive pairs, and other sample pairs as negative pairs, so as to pull positive pairs closer and push negative pairs further away:
\begin{equation}
\ell_{\mathrm{con}} = -\frac{1}{N} \sum_{i=1}^N 
\log \frac{
    \exp\left( {\mathbf{z}_i^u}^{\top}\mathbf{z}_i^v / \tau \right)
}{
    \sum_{j=1}^{N} \exp\left( {\mathbf{z}_i^u}^{\top} \mathbf{z}_j^v / \tau \right)
}.
\end{equation}
However, this hard segmentation strategy inevitably produces false negative pairs during the sample pair construction process, i.e., samples of the same type are regarded as negative pairs, which significantly reduces the quality of feature representation. 
To address this issue, recent methods~\cite{Lu0YPH024,GuoY00024} use similarity as weights to calculate additional positive pairs. While this approach improves the diversity of positive pairs and alleviates the issue of false negatives, the problem of false positive pairs remains unresolved. Moreover, the selection of positive and negative pairs is not reliable due to the influence of noisy representation. This instability ultimately degrades the robustness and generalization of MVC.
To address these limitations, we propose a reliability-aware noise contrastive learning that introduces a reliability graph to model inter-sample relationships and guide sample pair selection. The reliability graph is constructed as follows:
\begin{equation}
s_{ij}^{uv} = 
\begin{cases}
    \exp\left(-\dfrac{\|\mathbf{z}_i^u - \mathbf{z}_j^v\|^2}{\sigma}\right), & \text{if } i \ne j \\
    1, & \text{if } i = j
\end{cases}
\end{equation}
\begin{equation}
    \mathbf{A}^{uv} = \mathbf{SD}^{-1}.
\end{equation}
where, $\sigma$ is scaling factor, $s_{ij}^{uv}$ represents the similarity between the $i$-th sample of the $u$-th view and the $j$-th sample of the $v$-th view. $\mathbf{D}$ is degree matrix, $d_{ii} = \sum_j s_{ij}$, $a_{ij}$ represents the probability that sample $i$ and sample $j$ are judged as positive pairs. By converting the similarity information between samples into probability weights, similarity is used as a measure of the reliable relationship between samples to guide noise contrastive learning:
\begin{equation}
\ell_{\mathrm{ncon}}(u,v) = -\frac{1}{N}\sum_{i=1}^N\log\frac{\exp(\sum_{j=1}^{N}a^{uv}_{ij}\cdot{\mathbf{z}_i^u}^\top{\mathbf{z}_j^v}/\tau_c)}{\sum_{j=1}^{N}(1-a^{uv}_{ij})\cdot\exp({\mathbf{z}_i^u}^\top\mathbf{z}_j^v/\tau_c)}.
\end{equation}
Where $a_{ij}$ denote the reliability-aware indicator of a positive pair, where $(1 - a_{ij})$ naturally reflects a negative pair. $\tau_c$ denotes the temperature parameter in contrastive learning. By incorporating the reliability weights into contrastive training, the model can adaptively emphasize trustworthy sample pairs and downweight noisy or uncertain ones. This strategy not only improves the robustness of contrastive objectives under noisy conditions but also enhances the discriminative capacity of the learned representations by preserving reliable semantic structures. The final noise contrastive loss is defined as follows:
\begin{equation}
\ell_{\mathrm{ncon}} = \sum_{\substack{u,v \\ v \ne u}}^V\big(\ell_{\mathrm{ncon}}(u,v) + \ell_{\mathrm{ncon}}(u,u)\big).
\end{equation}
The noise contrastive loss consists of inter-view contrastive and intra-view contrastive.

\subsection{Dual-attention Imputation}
To reduce the impact of missing samples on model performance, we propose a method that combines intra-view and inter-view dual attention imputation. Specifically, for missing samples in $u$-th view, we use the corresponding samples in $v$-th view as queries, calculate their similarity with the observable samples in $u$-th view and the observable samples in $v$-th view. These observable samples are then treated as keys and values in an attention module to perform weighted imputation, thereby estimating the missing representations. The inter-view and intra-view attention weights are computed as follows:
\begin{align}
    \mathbf{S}_{\mathrm{inter}} = \mathrm{Softmax}\big((2{\mathbf{Z}^v_{[m^u]}}^\top\mathbf{Z}^u_{[\neg m^u]}-2)/\sigma\big),\\
    \mathbf{S}_{\mathrm{intra}} = \mathrm{Softmax}\big((2{\mathbf{Z}^v_{[m^u]}}^\top\mathbf{Z}^v_{[\neg m^v]}-2)/\sigma\big).
\end{align}
Where, $[m^u], [\neg m^u]$ represent the missing index and observable sample index of the $u$-th view respectively. $\mathbf{Z}^v_{[m^u]}$ indicates that the samples corresponding to $v$-th view are missing in $u$-th view. $\mathbf{Z}^u_{[\neg m^u]},\mathbf{Z}^v_{[\neg m^v]}$ represent the observable samples of views $u$ and $v$ respectively. $\mathbf{S}_{\mathrm{inter}}, \mathbf{S}_{\mathrm{intra}}$ represent the inter-view and intra-view attention relations respectively. The final imputation of missing samples in the $u$-th view is given by:
\begin{equation}
    \mathbf{Z}^u_{[m^u]} = \alpha\cdot\mathbf{Z}^u_{[\neg m^u]}\mathbf{S}_{\mathrm{inter}} + (1-\alpha)\cdot\mathbf{Z}^v_{[\neg m^v]}\mathbf{S}_{\mathrm{intra}}.
\end{equation}
This strategy not only retains the homogeneous structural information within the view, but also effectively integrates the heterogeneous completion information across views, thereby achieving a more robust estimation of missing features.

\subsection{Self-supervised Cluster Distillation}
In order to more effectively utilize the shared information to guide view-specific feature learning, we designed a self-supervised cluster distillation to constrain the distribution relationship between the fusion features and the view-specific features. Specifically, the view-specific features $\mathbf{Z}^v\in\mathbb{R}^{d\times N}$ are first concatenated to form a fusion feature matrix $\mathbf{Z}\in \mathbb{R}^{D\times N},D=Vd$. Then, the fused features are clustered with $k$-means to obtain pseudo labels and calculate $K$ cluster centers $\mathbf{C}\in\mathbb{R}^{D\times K}$:
\begin{equation}
    \mathbf{c}_k = \frac{1}{|\mathcal{M}_k|}\sum_{i\in \mathcal{M}_k}\mathbf{z}_i.
\end{equation}
where, $\mathcal{M}_k$ represents the feature set with pseudo label $k$. For each fused feature $\mathbf{z}_i$, the soft target distribution $\mathbf{Q}$ of cluster assignment is defined as follows:
\begin{equation}
    q_{ij} = \frac{\exp({\mathbf{z}_i}^\top\mathbf{c}_j/\tau_d)}{\sum_{k=1}^K\exp(\mathbf{z}_i^\top\mathbf{c}_k/\tau_d)}.
\end{equation}
Where, $\mathbf{q}_{ij}$ represents the probability that the $i$-th feature belongs to the $j$-th cluster. $\tau_d$ represents the temperature parameter, which is used to control the sharpness of the distribution. 

For each view, we introduce a trainable clustering layer, use randomly initialized parameters $\boldsymbol{\mu}^v$ as cluster centers, and calculate the predicted distribution $\mathbf{P}^v$ between view features and view cluster centers:
\begin{equation}
    p^v_{ij} = \frac{\exp({\mathbf{z}^v_i}^\top\boldsymbol{\mu}_j^v/\tau_d)}{\sum_{k=1}^K\exp({\mathbf{z}^v_i}^\top\boldsymbol{\mu}_k^v/\tau_d)}.
\end{equation}
Finally, the cluster distillation loss is defined as the sum of the Kullback-Leibler divergence between the target distribution $\mathbf{Q}$ and the prediction distribution $\mathbf{P}^v$ for each view:
\begin{equation}
    \ell_{\mathrm{dist}} = \sum_{v=1}^V\text{KL}(\mathbf{P}^v\parallel\mathbf{Q}).
\end{equation}

To summarize, the total loss of our method consists of cross-view reconstruction loss, noise contrastive loss, and distillation loss:
\begin{equation}
    \ell = \ell_{\mathrm{crec}} + \ell_{\mathrm{ncon}} + \ell_{\mathrm{dist}}.
\end{equation}

\section{Experiments}
\begin{table*}[htbp]
    \centering
    \setlength{\tabcolsep}{2pt}
    \caption{The clustering performance on five multi-view datasets at different noise ratios (The missing noise is set to the same ratio as the observation noise)}
    \begin{tabular}{clccccccccccccccc}
    \toprule
    \multirow{2}*{Ratio} & \multirow{2}*{Methods} & \multicolumn{3}{c}{Scene15} & \multicolumn{3}{c}{Caltech101} & \multicolumn{3}{c}{LandUse21} & \multicolumn{3}{c}{Reuters} & \multicolumn{3}{c}{NUS-WIDE} \\
    \cmidrule(lr){3-5}\cmidrule(lr){6-8}\cmidrule(lr){9-11}\cmidrule(lr){12-14}\cmidrule(lr){15-17}
    &  & ACC & NMI & ARI & ACC & NMI & ARI & ACC & NMI & ARI & ACC & NMI & ARI & ACC & NMI & ARI\\
    \midrule
    \multirow{9}*{0\%}
    & DCCAE~\cite{WangALB15} & 34.6 & 39.0 & 19.7 & 45.8 & 68.6 & 37.7 & 15.6 & 24.4 & 4.4 & 42.0 & 20.3 & 8.5 & 47.5 & 17.1 & 37.6\\
    & DSIMVC~\cite{TangL22} & 31.7 & 35.6 & 17.2 & 19.7 & 40.0 & 19.7 & 18.1 & 18.6 & 5.6 & 43.2 & 23.3 & 19.0 & 44.1 & 35.7 & 27.6\\
    & DIMVC~\cite{XuL0PMS022} & 35.5 & 36.4 & 18.1 & 38.7 & 56.7 & 17.1 & 26.7 & 32.4 & 12.5 & 48.2 & 22.8 & 19.4 & 54.0 & 43.3 & 33.8\\
    & DCP~\cite{LinGLBLP23} & 41.1 & 45.1 & 24.8 & 51.3 & 74.8 & 51.9& 26.2 & 32.7 & 13.5 & 36.2 & 18.9 & 4.8 & 53.3 & 42.4 & 28.6\\
    & SURE~\cite{Yang00B0023} & 41.0 & 43.2 & 25.0 & 43.8 & 70.1 & 29.5 & 25.1 & 28.3 & 10.9 & 52.1 & 36.9 & 26.6 & 57.4 & 44.8 & \underline{38.3} \\
    & ProImp~\cite{Li0Y0P023} & 43.6 & 45.0 & 26.8 & 37.6 & 67.0 & 25.0 & 23.7 & 27.9 & 10.8 & 56.5 & 39.4 & 32.8 & 52.2 & 43.6 & 31.2\\
    & DIVIDE~\cite{Lu0YPH024} & \underline{49.1} & \underline{48.7} & \underline{31.6} & 62.2 & 83.0 & \underline{50.5} & \underline{32.3} & \underline{39.7} & \underline{18.1} & \underline{59.3} & \underline{39.5} & 29.0 & 45.1 & 30.9 & 19.4 \\
    & CANDY~\cite{GuoY00024} & 42.0 & 41.6 & 24.7 & \underline{67.3} & \textbf{83.8} & \textbf{60.0} & 30.6 & 36.5 & 16.2 & 57.7 & 30.8 & \underline{37.1} & \underline{62.1} & \underline{49.0} & 37.0\\
    & PMIMC~\cite{YuanSZWYYR25} & 32.8 & 36.8 & 18.4 & 45.0 & 71.1 & 35.4 & 25.7 & 33.7 & 11.6 & 51.3 & 31.6 & 25.3 & 39.9 & 34.7 & 21.6\\
    & Ours & \textbf{49.7} & \textbf{49.6} & \textbf{32.6} & \textbf{67.9} & \underline{83.1} & 49.8 & \textbf{32.6} & \textbf{40.3} & \textbf{18.7} & \textbf{65.3} & \textbf{43.8} & \textbf{39.5} & \textbf{65.9} & \textbf{52.6} & \textbf{47.0}\\
    \midrule
    \multirow{9}*{20\%}
    & DCCAE~\cite{WangALB15} & 30.0 & 28.3 & 13.6 & 45.6 & 71.1 & 32.2 & 21.2 & 21.6 & 7.3 & 42.8 & 18.4 & 15.0 & 43.9 & 31.1 & 22.5\\
    & DSIMVC~\cite{TangL22} & 27.4 & 27.1 & 14.1 & 18.7 & 29.7 & 12.4 & 15.1 & 14.5 & 3.7 & 36.8 & 16.4 & 14.0 & 18.5 & 13.6 & 7.2\\
    & DIMVC~\cite{XuL0PMS022} & 30.4 & 30.2 & 14.1& 40.9 & 56.9 & 35.3 & 20.6 & 25.4 & 7.4 & 48.1 &  23.4 & 18.4 & 41.8 & 30.9 & 22.4\\
    & DCP~\cite{LinGLBLP23} & 38.5 & 40.2 & 22.8 & 44.0 & 66.3 & 58.9 & 26.0 & 28.4 & 12.1 & 38.8 & 21.6 & 7.8 & 47.4 & 41.2 & 25.8\\
    & SURE~\cite{Yang00B0023} & 40.1 & 38.8 & 22.0 & 52.1 & 74.8 & 39.1 & 27.3 & 31.0 & 13.8 & 53.3 & 36.9 & 28.5 & 54.3 & 41.9 & 35.5\\
    & ProImp~\cite{Li0Y0P023} & 39.8 & \underline{42.4} & \underline{24.2} & 39.9 & 68.0 & 27.3 & 22.4 & 26.3 & 9.8 & 54.4 & \underline{39.3} & 30.8 & 55.8 & 44.0 & \underline{38.0}\\
    & DIVIDE~\cite{Lu0YPH024} & \underline{41.1} & 41.0 & 24.1 & 67.6 & 83.1 & 60.9 & \underline{31.4} & \underline{37.7} & \underline{16.9} & 57.0 & 37.6 & 31.8 & 59.3 & \underline{44.6} & 37.7\\
    & CANDY~\cite{GuoY00024} & 40.7 & 38.1 & 21.6 & \underline{68.6} & \underline{83.8} & \underline{62.7} & 30.7 & 33.3 & 15.1 & \underline{60.3} & 38.0 & \underline{33.9} & \underline{60.0} & 37.5 & 33.6\\
    & PMIMC~\cite{YuanSZWYYR25} & 33.3 & 30.9 & 17.3 & 37.9 & 66.2 & 30.8 & 18.4 & 23.3 & 6.2 & 46.3 & 24.3 & 19.4 & 26.4 & 15.0 & 7.5\\
    & Ours & \textbf{45.2} & \textbf{42.9} & \textbf{26.9} & \textbf{69.3} & \textbf{83.9} & \textbf{64.4} & \textbf{31.7} & \textbf{38.8} & \textbf{18.3} & \textbf{65.1} & \textbf{42.0} & \textbf{39.4} & \textbf{62.6} & \textbf{47.4} & \textbf{40.8}\\
    \midrule
    \multirow{9}*{50\%}
    & DCCAE~\cite{WangALB15} & 19.6 & 17.5 & 6.5 & 35.0 & 65.6 & 27.7 & 13.4 & 12.2 & 2.6 & 33.1 & 9.9 & 12.0 & 33.8 & 23.7 & 14.6\\
    & DSIMVC~\cite{TangL22} & 26.9 & 26.1 & 13.8 & 16.0 & 24.0 & 4.8 & 11.5 & 9.1 & 2.6 & 34.6 & 14.3 & 12.6 & 12.2 & 1.5 & 0.1\\
    & DIMVC~\cite{XuL0PMS022} & 23.2 & 21.9 & 8.9 & 27.9 & 47.5 & 15.3 & 13.6 & 13.6 & 2.7 & 44.6 & 18.7 & 15.4 & 33.3 & 20.1 & 14.2\\
    & DCP~\cite{LinGLBLP23} & 31.4 & 31.3 & 16.8 & 48.3 & 70.4 & 59.6 & 21.5 & 22.3 & 8.4 & 37.0 & 21.6 & 5.8 & 17.7 & 13.5 & 8.4\\
    & SURE~\cite{Yang00B0023} & 35.4 & \underline{34.7} & 18.2 & 52.2 & 75.1 & 40.9 & \textbf{26.1} & \textbf{28.1} & \textbf{12.6} & 52.2 & 36.0 & 28.7 & 52.2 & 39.0 & \underline{32.7}\\
    & ProImp~\cite{Li0Y0P023} & \underline{36.8} & 34.4 & \underline{19.2} & 34.7 & 63.7 & 24.5 & 19.0 & 18.2 & 5.8 & 41.7 & 16.7 & 14.9 & 50.2 & 33.9 & 25.2\\
    & DIVIDE~\cite{Lu0YPH024} & 32.5 & 33.9 & 17.5 & 64.3 & \underline{83.2} & 59.9 & 24.0 & \underline{27.1} & 9.5 & 54.8 & 36.0 & \underline{31.3} & \underline{54.7} & \textbf{40.3} & 31.1\\
    & CANDY~\cite{GuoY00024} & 34.9 & 32.7 & 16.9 & \underline{67.7} & \underline{83.2} & \underline{62.7} & 23.8 & 26.3 & 10.1 & \underline{54.9} & \underline{38.4} & 30.0 & 48.2 & 33.9 & 25.7\\
    & PMIMC~\cite{YuanSZWYYR25} & 30.7 & 29.1 & 15.2 & 41.2 & 68.5 & 31.7 & 18.0 & 21.9 & 5.5 & 43.2 & 17.8 & 17.3 & 21.3 & 8.2 & 4.8\\
    & Ours & \textbf{42.1} & \textbf{35.3} & \textbf{21.5} & \textbf{73.6} & \textbf{85.9} & \textbf{73.9} & \underline{25.9} & 24.4 & \underline{10.8} & \textbf{61.7} & \textbf{36.6} & \textbf{34.6} & \textbf{55.9} & \underline{39.2} & \textbf{32.8}\\
    \midrule
    \multirow{9}*{80\%}
    & DCCAE~\cite{WangALB15} & 16.6 & 15.6 & 5.1 & 26.5 & 58.2 & 24.0 & 15.7 & 13.6 & 3.5 & 29.5 & 8.5 & 7.4 & 20.7 & 14.0 & 6.8\\
    & DSIMVC~\cite{TangL22} & 18.9 & 16.6 & 9.6 & 16.1 & 23.5 & 5.3 & 11.8 & 7.8 & 1.5 & 32.5 & 9.7 & 9.1 & 12.6 & 1.2 & 0.1\\
    & DIMVC~\cite{XuL0PMS022} & 14.9 & 9.4 & 2.3& 30.7 & 53.0 & 27.5 & 14.5 & 13.1 & 2.7 & 38.7 &  13.0 & 8.3 & 17.8 & 6.2 & 1.9\\
    & DCP~\cite{LinGLBLP23} & 20.1 & 20.0 & 7.2 & 39.7 & 61.3 & \underline{50.4} & 17.1 & 16.0 & 3.2 & 39.5 & 25.5 & 14.9 & 12.0 & 1.0 & 0.1\\
    & SURE~\cite{Yang00B0023} & \textbf{36.0} & \textbf{37.7} & \textbf{20.1} & 50.4 & 72.8 & 41.8 & 25.9 & 27.8 & 12.1 & 46.3 & 31.2 & 23.5 & 40.3 & {32.2} & {22.5}\\
    & ProImp~\cite{Li0Y0P023} & 25.8 & 21.4 & 10.2 & 34.3 & 63.7 & 25.4 & 15.5 & 14.4 & 4.0 & 26.6 & 5.5 & 4.4 & 31.8 & 19.0 & 12.4\\
    & DIVIDE~\cite{Lu0YPH024} & 27.6 & 29.3 & 15.5 & \underline{61.8} & \underline{80.5} & 49.2 & 28.3 & {34.1} & \underline{13.6} & \underline{53.1} & \textbf{34.3} & \textbf{30.2} & \underline{48.7} & \underline{33.5} & \underline{25.2}\\
    & CANDY~\cite{GuoY00024} & 26.5 & \underline{29.9} & 14.1 & 61.5 & 80.1 & 48.6 & \underline{29.0} & \underline{35.3} & {13.5} & 47.2 & 25.8 & 22.9 & {44.0} & 24.4 & 20.3\\
    & PMIMC~\cite{YuanSZWYYR25} & 27.6 & 23.4 & 12.0 & 39.6 & 67.6 & 31.4 & 15.9 & 19.4 & 4.9 & 44.5 & 17.6 & 15.2 & 20.4 & 6.8 & 3.7\\
    & Ours & \underline{32.4} & {27.5} & \underline{16.8} & \textbf{67.8} & \textbf{82.9} & \textbf{57.1} & \textbf{29.2} & \textbf{36.1} & \textbf{15.9} & \textbf{55.5} & \underline{31.7} & \underline{27.6} & \textbf{50.6} & \textbf{35.0} & \textbf{30.8}\\
    \bottomrule
    \end{tabular}
    \label{tab_comp}
\end{table*}

\begin{figure*}
    \centering
    \subfloat[Scene15]{\includegraphics[width=0.2\linewidth]{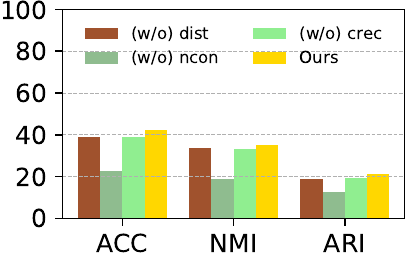}}
    \subfloat[Caltech101]{\includegraphics[width=0.2\linewidth]{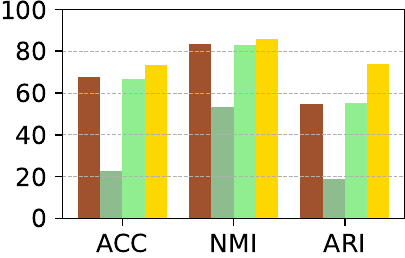}}
    \subfloat[LandUse21]{\includegraphics[width=0.2\linewidth]{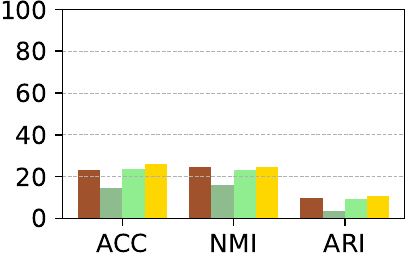}}
    \subfloat[Reuters]{\includegraphics[width=0.2\linewidth]{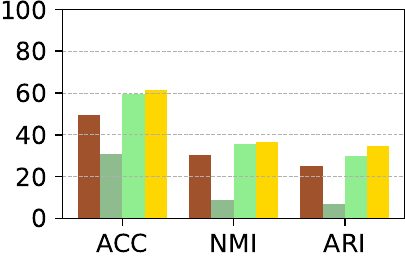}}
    \subfloat[NUS-WIDE]{\includegraphics[width=0.2\linewidth]{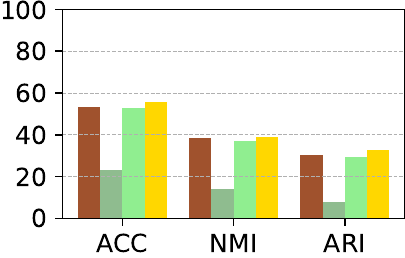}}
    \caption{Ablation study on five dataset with 50\% noise ratio}
    \label{fig_ablation}
\end{figure*}

\subsection{Datasets and Evaluation Metrics}
\begin{itemize}
   \item \textbf{Scene-15}~\cite{LiPT05} includes 4,485 images across 15 categories. We employ PHOG and GIST as two distinct views following~\cite{Yang00B0023}.
   \item \textbf{Caltech-101}~\cite{LiNHH15} consists 8,677 images collected from 101 classes. We use two kinds of deep features extracted by the DECAF and VGG19 neural networks as two views following~\cite{HanZFZ21}.
   \item \textbf{LandUse-21}~\cite{YangN10} contains 2,100 satellite imagery samples in 21 categories. We employ the PHOG and LBP features as two views following~\cite{LinGLBLP23}.
   \item \textbf{Reuters}~\cite{AminiUG09} is a repository of news content in multiple languages with 18,758 samples. Following~\cite{HuangZ0ZZL19}, we transform the texts into a 10-dimensional latent space with a conventional autoencoder and use English and French as two different views.
   \item \textbf{NUS-WIDE}~\cite{HuZPL19} includes 9,000 images paired with their respective captions from 10 classes. We adopt a VGG19 neural network for the extraction of visual features, and a Sentence CNN to extract the text features by following~\cite{ZhenHWP19}.
\end{itemize}
There are three metrics are widely used in MVC, including clustering accuracy (ACC), normalized mutual information (NMI) and adjusted rand index (ARI). Higher values of these indicate better clustering performance.

\subsection{Implementation Details}
The encoder is a four-layer MLP, and the decoder adopts a symmetric architecture per view. The latent feature dimension $d$ is set to 128, and the clustering layer is $d-K$.
We use the Adam optimizer with a learning rate of $2 \times 10^{-3}$ and a batch size of 1024. Temperature parameters $\tau_c$ and $\tau_d$ are both set to 0.5, the scaling factor $\sigma$ is 0.07, and the imputation weight $\alpha$ is 0.5. Unless otherwise noted, missing noise ratio $\eta_m$ and observation noise ratio $\eta_n$ are both set to 0.5.
All experiments are repeated multiple times, and average results are reported. Training is performed on a desktop with an Intel i7-12700KF CPU, NVIDIA RTX 3080Ti GPU, and 32GB RAM using the PyTorch framework. For all baselines, we use publicly available source code and official configurations to ensure fair comparisons.

\subsection{Comparisons with State of the Arts}
We compared the proposed method with nine representative DMVC methods, including DCCAE~\cite{WangALB15}, DSIMVC~\cite{TangL22}, DIMVC~\cite{XuL0PMS022}, DCP~\cite{LinGLBLP23}, SURE~\cite{Yang00B0023}, ProImp~\cite{Li0Y0P023}, DIVIDE~\cite{Lu0YPH024}, CANDY~\cite{GuoY00024} and PMIMC~\cite{YuanSZWYYR25}.
As shown in Table~\ref{tab_comp}, our method maintains stable performance in both the noise-free setting and under varying ratios of multi-source noise. These results indicate the effectiveness and robustness of the framework in handling missing noise and observation noise.
It is worth noting that DIVIDE and CANDY reduce the impact of false negative pairs by adjusting the contrastive weights by introducing high-order graph structures. However, high-order graphs may amplify noise through propagation. In addition, the presence of false positive pairs can further impair the clustering performance. In contrast, our method leverages a reliability-aware graph to guide contrastive learning under noise, explicitly accounting for the effects of both false negative and false positive pairs. This design effectively enhances the discriminative ability of the learned features and improves the robustness of the model in noisy environments.

\begin{figure}
    \centering
    \subfloat[Clean 0]{\includegraphics[width=0.25\linewidth]{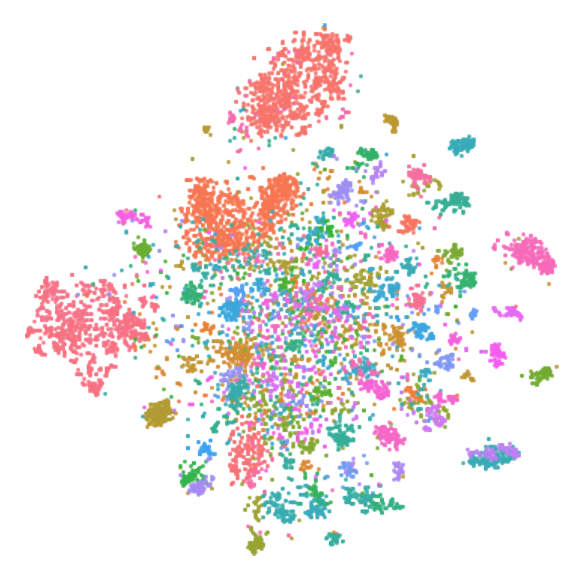}}
    \subfloat[Noisy 0]{\includegraphics[width=0.25\linewidth]{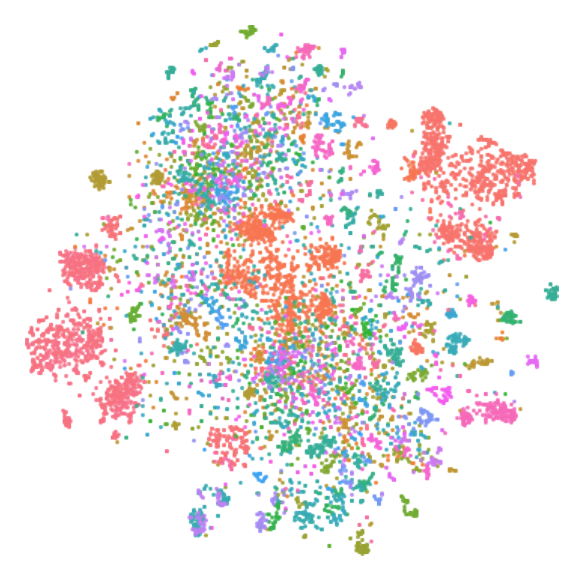}}
    \subfloat[Clean 99]{\includegraphics[width=0.25\linewidth]{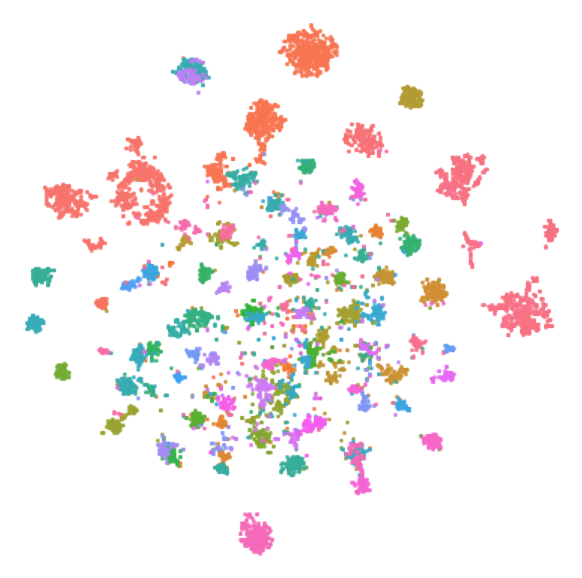}}
    \subfloat[Noisy 99]{\includegraphics[width=0.25\linewidth]{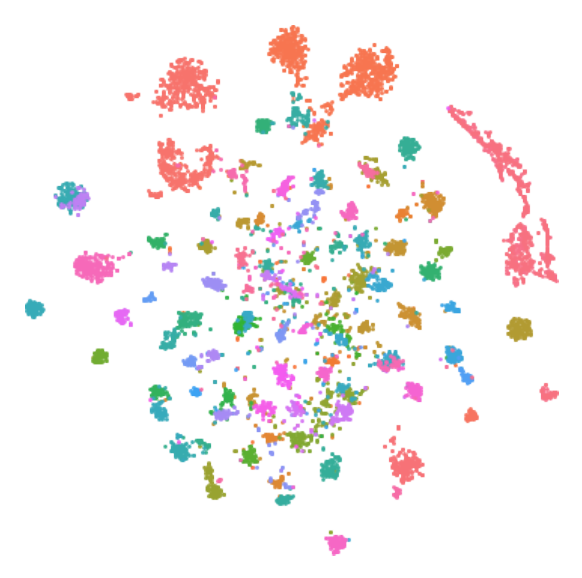}}
    \caption{Visualization of the training process on Caltech101. 0 and 99 represent the epoch of training.}
    \label{fig_visual}
\end{figure}
\subsection{Visualization Analysis}
To illustrate the data distribution under noisy conditions, we use $t$-SNE to visualize and compare the training process of the Caltech101 dataset under clean views and 50\% multi-source noise.
As shown in Fig.~\ref{fig_visual}, before training, the distribution of noisy views is noticeably more scattered than that of the clean views. After 100 epochs of training, the distributions become more aligned.
Under noisy conditions, the learned representations have more compact intra-cluster structure and clearer inter-cluster separation compared to the initial state, demonstrating that the our method maintains good clustering performance even in multi-source noise scenarios.

\subsection{Ablation Study}
\subsubsection{Effectiveness of design modules:}
To assess the contribution of each component to the overall performance, we conduct a systematic ablation study. As illustrated in the Fig.~\ref{fig_ablation}, "(w/o) dist", "(w/o) crec" and "(w/o) ncon", and "ours" denote the model variants obtained by removing the self-supervised distillation module, cross-view reconstruction module, noise contrastive learning module, and the complete model, respectively.
We evaluate all variants on five benchmark datasets using three standard clustering metrics under 50\% noise ratio. The results demonstrate that each module contributes positively to performance improvement. In particular, the noise contrastive learning module plays a critical role by enhancing multi-view feature consistency and significantly mitigating the influence of observation noise. Additionally, the cross-view reconstruction mechanism improves robustness by suppressing noise at the data level, while the self-supervised distillation module further refines feature representations and improves clustering performance.

\begin{table}
    \centering
    \caption{Effectiveness of noise contrastive learning}
    \begin{tabular}{lcccccc}
    \toprule
     & \multicolumn{3}{c}{Scene15} & \multicolumn{3}{c}{Caltech101}\\
    \cmidrule(lr){2-4}\cmidrule(lr){5-7}
     & ACC & NMI & ARI & ACC & NMI & ARI \\
    \midrule
    Con & 40.8 & 34.1 & 20.4 & 58.7 & 81.8 & 39.9 \\
    FN con & 41.7 & 34.4 & 21.1 & 65.7 & 82.8 & 51.1 \\
    Ours & \textbf{42.1} & \textbf{35.3} & \textbf{21.5} & \textbf{73.6} & \textbf{85.9} & \textbf{73.9} \\
    \bottomrule
    \end{tabular}
    \label{tab_contrastive}
\end{table}
\subsubsection{Effectiveness of noise contrastive learning:}
To analyze the effectiveness of the proposed reliability-aware noise contrastive learning, we compare it against two baselines: the traditional contrastive learning method (Con) and the false negative pair-aware contrastive method (FN Con) on two datasets under 50\% noise ratio.
As shown in Table~\ref{tab_contrastive}, both the traditional and FN contrastive methods fail to perform effectively in noisy environments. 
In contrast, our method dynamically adjusts the weights of sample pairs based on their estimated reliability. This strategy not only mitigates the effects of both false negative and false positive pairs, but also reduces the impact of noise on the constructions, thereby enhancing the robustness and discriminability of the learned representations.

\begin{table}
    \centering
    \setlength{\tabcolsep}{5pt}
    \caption{Effectiveness of imputation}
    \begin{tabular}{lcccccc}
    \toprule
     & \multicolumn{3}{c}{Reuters} & \multicolumn{3}{c}{NUS-WIDE} \\
    \cmidrule(lr){2-4}\cmidrule(lr){5-7}
     & ACC & NMI & ARI & ACC & NMI & ARI\\
    \midrule
    Directly & 39.9 & 21.1 & 13.3 & 34.4 & 23.8 & 9.4\\
    Prototype & 51.8 & 26.7 & 22.6 & 52.0 & 35.3 & 29.2\\
    $k$-nn & 57.6 & 32.6 & 32.1 & 43.0 & 31.0 & 13.9\\
    Ours & \textbf{61.7} & \textbf{36.6} & \textbf{34.6} & \textbf{55.9} & \textbf{39.2} & \textbf{32.8}\\
    \bottomrule
    \end{tabular}
    \label{tab_imputation}
\end{table}
\subsubsection{Effectiveness of imputation:}
To evaluate the effectiveness of the proposed missing view imputation method, we compare it with three commonly used imputation strategies on two benchmark datasets under 50\% noise ratio: direct imputation, prototype imputation, and $k$-nn imputation. As shown in Table~\ref{tab_imputation}, direct imputation replaces the missing view with features from the corresponding view, which may lead to feature homogenization and degradation of multi-view representation quality. Prototype imputation uses the cluster centroid of the observable view for imputation, which introduces discriminative information but compromises view-specific information. $k$-nn imputation fills in missing data using neighboring features from the same view, preserving view-specific information but ignoring cross-view interactions. In contrast, our method effectively integrates information from multiple views while  preserving view-specific features, leading to improved imputation quality and clustering performance.

\subsection{Hyper-parameter Analysis}
\begin{figure}
    \centering
    \subfloat[ACC]{\includegraphics[width=0.5\linewidth]{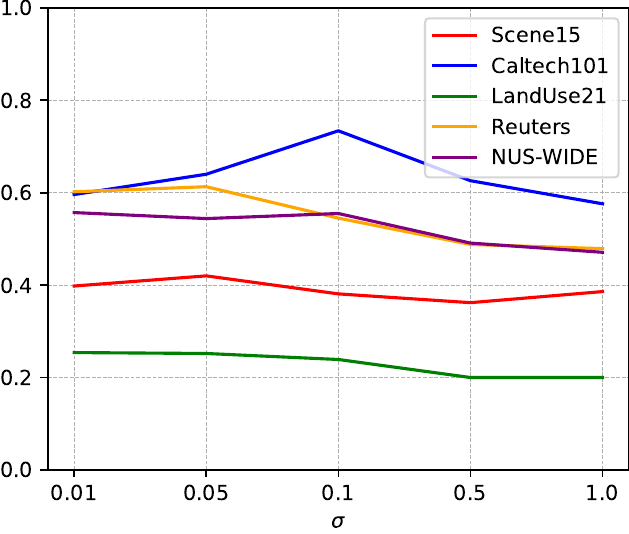}}
    \subfloat[NMI]{\includegraphics[width=0.5\linewidth]{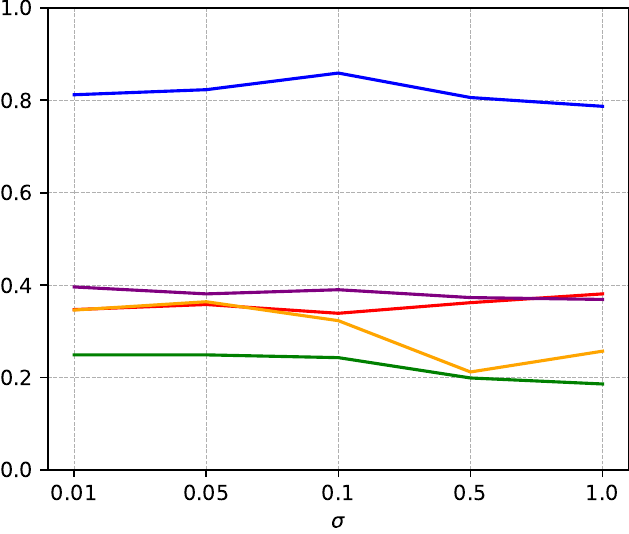}}
    \caption{The impact of $\sigma$ on performance.}
    \label{fig_sigma}
\end{figure}
\subsubsection{Sensitivity analysis of hyperparameter $\sigma$:}
To evaluate the effect of the graph construction scaling factor $\sigma$ on model performance, we conduct experiments on five datasets under 50\% noise ratio with $\sigma \in \{0.01, 0.05, 0.1, 0.5, 1.0\}$. As shown in Fig.~\ref{fig_sigma}, the model maintains generally stable performance across different values. Notably, performance improves consistently when $\sigma \leq 0.1$, indicating that a smaller $\sigma$ facilitates better separation of positive and negative pairs and reduces the impact of noisy samples.

\begin{figure}
    \centering
    \subfloat[Reuters ACC]{\includegraphics[width=0.45\linewidth]{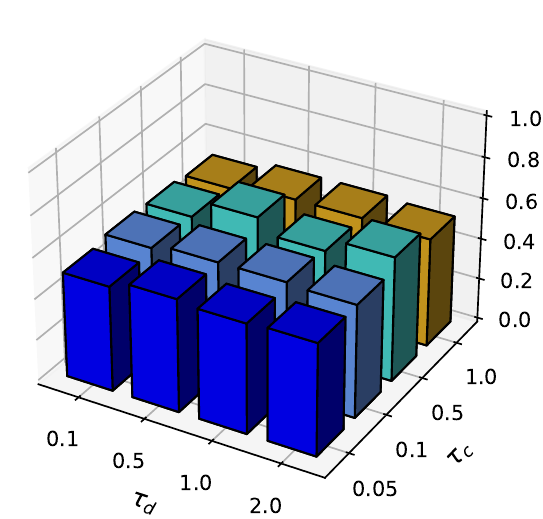}}
    \subfloat[Reuters NMI]{\includegraphics[width=0.45\linewidth]{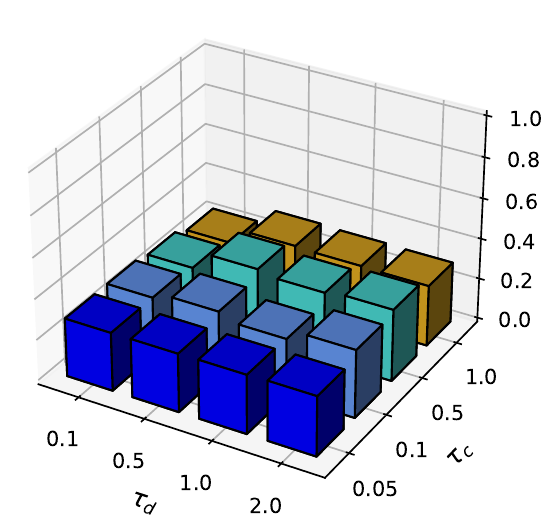}}\\
    \subfloat[NUS-WIDE ACC]{\includegraphics[width=0.45\linewidth]{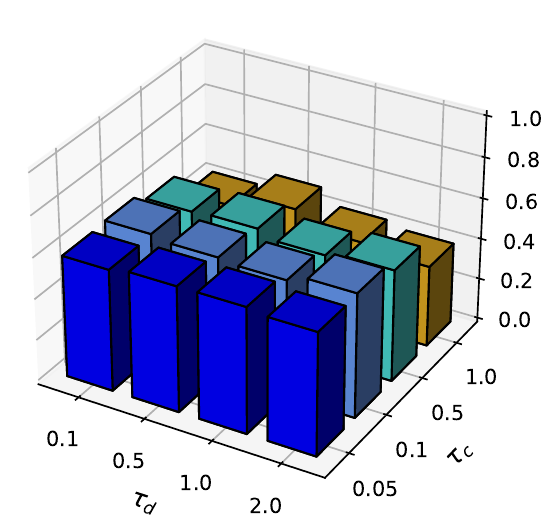}}
    \subfloat[NUS-WIDE NMI]{\includegraphics[width=0.45\linewidth]{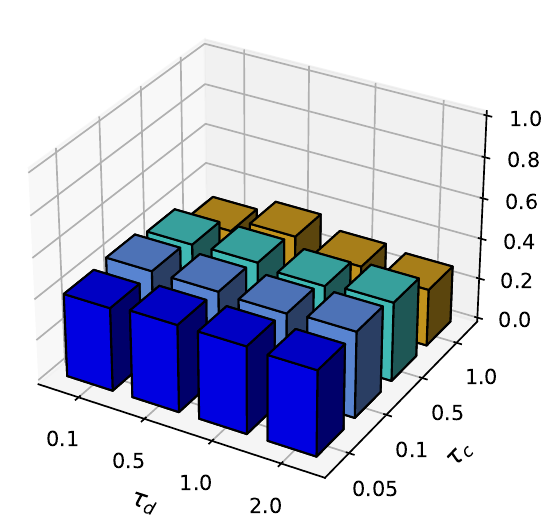}}
    \caption{The impact of $\tau_c,\tau_d$ on performance.}
    \label{fig_tau}
\end{figure}
\subsubsection{Sensitivity analysis of hyperparameter $\tau_c, \tau_d$:}
We further analyzed the impact of the contrastive temperature parameter $\tau_c$ and the distillation temperature parameter $\tau_d$ on the performance on the Reuters and NUS-WIDE datasets under 50\% noise ratio. As shown in Fig.~\ref{fig_tau}, on the Reuters data, when $\tau_c \in [0.05,0.5]$ and $\tau_d\in[0.5,1.0]$, the model performs best; on NUS-WIDE, the better performance corresponds to $\tau_c\in[0.05, 0.1]$ and $\tau_d\in[0.1,0.5]$. This shows that smaller contrastive and distillation temperatures help sharpen the similarity between feature distributions and the discriminability of cluster-level distributions, thereby improving clustering performance.

\section{Conclusion}
In this paper, we proposed a reliability-aware contrastive framework for robust multi-view clustering under multi-source noise. The framework is designed to handle both observation noise and view missingness through integrated mechanisms. Extensive experiments on five benchmarks confirm that our method achieves superior performance and maintains robustness across varying noise levels, demonstrating its effectiveness in noisy real-world scenarios.

\section{Acknowledgments}
This work was supported by the National Natural Science Foundation of China under grant 92470202 and Frontier Technologies R\&D Program of Jiangsu under grant BF2024070.

\bibliography{aaai2026}

@inproceedings{LiPT05,
  author       = {Li Fei{-}Fei and
                  Pietro Perona},
  title        = {A Bayesian Hierarchical Model for Learning Natural Scene Categories},
  booktitle    = {2005 {IEEE} Computer Society Conference on Computer Vision and Pattern
                  Recognition, San Diego, CA},
  pages        = {524--531},
  publisher    = {{IEEE} Computer Society},
  year         = {2005}
}

@article{Yang00B0023,
  author       = {Mouxing Yang and
                  Yunfan Li and
                  Peng Hu and
                  Jinfeng Bai and
                  Jiancheng Lv and
                  Xi Peng},
  title        = {Robust Multi-View Clustering With Incomplete Information},
  journal      = {{IEEE} Trans. Pattern Anal. Mach. Intell.},
  volume       = {45},
  number       = {1},
  pages        = {1055--1069},
  year         = {2023}
}

@inproceedings{LiNHH15,
  author       = {Yeqing Li and
                  Feiping Nie and
                  Heng Huang and
                  Junzhou Huang},
  title        = {Large-Scale Multi-View Spectral Clustering via Bipartite Graph},
  booktitle    = {Proceedings of the Twenty-Ninth {AAAI} Conference on Artificial Intelligence, Austin, Texas},
  pages        = {2750--2756},
  publisher    = {{AAAI} Press},
  year         = {2015}
}

@inproceedings{HanZFZ21,
  author       = {Zongbo Han and
                  Changqing Zhang and
                  Huazhu Fu and
                  Joey Tianyi Zhou},
  title        = {Trusted Multi-View Classification},
  booktitle    = {9th International Conference on Learning Representations,
                  Virtual Event, Austria, May 3-7, 2021},
  publisher    = {OpenReview.net},
  year         = {2021}
}

@inproceedings{YangN10,
  author       = {Yi Yang and
                  Shawn D. Newsam},
  title        = {Bag-of-visual-words and spatial extensions for land-use classification},
  booktitle    = {18th International Symposium on Advances in Geographic
                  Information Systems, San Jose,
                  CA, Proceedings},
  pages        = {270--279},
  publisher    = {{ACM}},
  year         = {2010}
}

@article{LinGLBLP23,
  author       = {Yijie Lin and
                  Yuanbiao Gou and
                  Xiaotian Liu and
                  Jinfeng Bai and
                  Jiancheng Lv and
                  Xi Peng},
  title        = {Dual Contrastive Prediction for Incomplete Multi-View Representation Learning},
  journal      = {{IEEE} Trans. Pattern Anal. Mach. Intell.},
  volume       = {45},
  number       = {4},
  pages        = {4447--4461},
  year         = {2023}
}

@inproceedings{AminiUG09,
  author       = {Massih{-}Reza Amini and
                  Nicolas Usunier and
                  Cyril Goutte},
  title        = {Learning from Multiple Partially Observed Views - an Application to
                  Multilingual Text Categorization},
  booktitle    = {Advances in Neural Information Processing Systems 22, Vancouver, British Columbia,
                  Canada},
  pages        = {28--36},
  publisher    = {Curran Associates, Inc.},
  year         = {2009}
}

@inproceedings{HuangZ0ZZL19,
  author       = {Zhenyu Huang and
                  Joey Tianyi Zhou and
                  Xi Peng and
                  Changqing Zhang and
                  Hongyuan Zhu and
                  Jiancheng Lv},
  title        = {Multi-view Spectral Clustering Network},
  booktitle    = {Proceedings of the Twenty-Eighth International Joint Conference on
                  Artificial Intelligence, Macao},
  pages        = {2563--2569},
  publisher    = {ijcai.org},
  year         = {2019}
}

@inproceedings{HuZPL19,
  author       = {Peng Hu and
                  Liangli Zhen and
                  Dezhong Peng and
                  Pei Liu},
  title        = {Scalable Deep Multimodal Learning for Cross-Modal Retrieval},
  booktitle    = {Proceedings of the 42nd International {ACM} {SIGIR} Conference on
                  Research and Development in Information Retrieval, Paris,
                  France},
  pages        = {635--644},
  publisher    = {{ACM}},
  year         = {2019}
}

@inproceedings{ZhenHWP19,
  author       = {Liangli Zhen and
                  Peng Hu and
                  Xu Wang and
                  Dezhong Peng},
  title        = {Deep Supervised Cross-Modal Retrieval},
  booktitle    = {{IEEE} Conference on Computer Vision and Pattern Recognition, Long Beach, CA},
  pages        = {10394--10403},
  publisher    = {Computer Vision Foundation / {IEEE}},
  year         = {2019}
}

@inproceedings{WangALB15,
  author       = {Weiran Wang and
                  Raman Arora and
                  Karen Livescu and
                  Jeff A. Bilmes},
  title        = {On Deep Multi-View Representation Learning},
  booktitle    = {Proceedings of the 32nd International Conference on Machine Learning,
                 Lille, France},
  series       = {{JMLR} Workshop and Conference Proceedings},
  volume       = {37},
  pages        = {1083--1092},
  publisher    = {JMLR.org},
  year         = {2015}
}

@inproceedings{Lu0YPH024,
  author       = {Yiding Lu and
                  Yijie Lin and
                  Mouxing Yang and
                  Dezhong Peng and
                  Peng Hu and
                  Xi Peng},
  title        = {Decoupled Contrastive Multi-View Clustering with High-Order Random Walks},
  booktitle    = {Thirty-Eighth Conference on Artificial Intelligence, Vancouver, Canada},
  pages        = {14193--14201},
  publisher    = {{AAAI} Press},
  year         = {2024}
}

@inproceedings{GuoY00024,
  author       = {Ruiming Guo and
                  Mouxing Yang and
                  Yijie Lin and
                  Xi Peng and
                  Peng Hu},
  title        = {Robust Contrastive Multi-view Clustering against Dual Noisy Correspondence},
  booktitle    = {Advances in Neural Information Processing Systems 38, Vancouver,
                  BC, Canada},
  year         = {2024}
}

@inproceedings{Li0Y0P023,
  author       = {Haobin Li and
                  Yunfan Li and
                  Mouxing Yang and
                  Peng Hu and
                  Dezhong Peng and
                  Xi Peng},
  title        = {Incomplete Multi-view Clustering via Prototype-based Imputation},
  booktitle    = {Proceedings of the Thirty-Second International Joint Conference on
                  Artificial Intelligence, Macao, SAR},
  pages        = {3911--3919},
  publisher    = {ijcai.org},
  year         = {2023}
}

@inproceedings{TangL22,
  author       = {Huayi Tang and
                  Yong Liu},
  title        = {Deep Safe Incomplete Multi-view Clustering: Theorem and Algorithm},
  booktitle    = {International Conference on Machine Learning, Baltimore, Maryland},
  series       = {Proceedings of Machine Learning Research},
  volume       = {162},
  pages        = {21090--21110},
  publisher    = {{PMLR}},
  year         = {2022}
}

@article{YuanSZWYYR25,
  author       = {Honglin Yuan and
                  Yuan Sun and
                  Fei Zhou and
                  Jing Wen and
                  Shihua Yuan and
                  Xiaojian You and
                  Zhenwen Ren},
  title        = {Prototype Matching Learning for Incomplete Multi-View Clustering},
  journal      = {{IEEE} Trans. Image Process.},
  volume       = {34},
  pages        = {828--841},
  year         = {2025}
}

@inproceedings{li2019deep,
  title={Deep Adversarial Multi-view Clustering Network.},
  author={Li, Zhaoyang and Wang, Qianqian and Tao, Zhiqiang and Gao, Quanxue and Yang, Zhaohua and others},
  booktitle={IJCAI},
  volume={2},
  number={3},
  pages={4},
  year={2019}
}

@inproceedings{xu2019adversarial,
  title={Adversarial incomplete multi-view clustering.},
  author={Xu, Cai and Guan, Ziyu and Zhao, Wei and Wu, Hongchang and Niu, Yunfei and Ling, Beilei},
  booktitle={IJCAI},
  volume={7},
  pages={3933--3939},
  year={2019}
}

@inproceedings{xu2021multi,
  title={Multi-VAE: Learning disentangled view-common and view-peculiar visual representations for multi-view clustering},
  author={Xu, Jie and Ren, Yazhou and Tang, Huayi and Pu, Xiaorong and Zhu, Xiaofeng and Zeng, Ming and He, Lifang},
  booktitle={Proceedings of the IEEE/CVF international conference on computer vision},
  pages={9234--9243},
  year={2021}
}

@article{abavisani2018deep,
  title={Deep multimodal subspace clustering networks},
  author={Abavisani, Mahdi and Patel, Vishal M},
  journal={IEEE Journal of Selected Topics in Signal Processing},
  volume={12},
  number={6},
  pages={1601--1614},
  year={2018},
  publisher={IEEE}
}

@inproceedings{TangTWFW18,
  author       = {Xiaoliang Tang and
                  Xuan Tang and
                  Wanli Wang and
                  Li Fang and
                  Xian Wei},
  title        = {Deep Multi-view Sparse Subspace Clustering},
  booktitle    = {Proceedings of the {VII} International Conference on Network, Communication
                  and Computing, Taipei City, Taiwan},
  pages        = {115--119},
  year         = {2018}
}

@inproceedings{cheng2021multi,
  title={Multi-view attribute graph convolution networks for clustering},
  author={Cheng, Jiafeng and Wang, Qianqian and Tao, Zhiqiang and Xie, Deyan and Gao, Quanxue},
  booktitle={Proceedings of the twenty-ninth international conference on international joint conferences on artificial intelligence},
  pages={2973--2979},
  year={2021}
}

@inproceedings{0001PCZCP024,
  author       = {Yazhou Ren and
                  Jingyu Pu and
                  Chenhang Cui and
                  Yan Zheng and
                  Xinyue Chen and
                  Xiaorong Pu and
                  Lifang He},
  title        = {Dynamic Weighted Graph Fusion for Deep Multi-View Clustering},
  booktitle    = {Proceedings of the Thirty-Third International Joint Conference on
                  Artificial Intelligence},
  pages        = {4842--4850},
  publisher    = {ijcai.org},
  year         = {2024}
}

@article{AsadAMJAYL25,
  author       = {Mujtaba Asad and
                  Waqar Azeem and
                  Aftab Ahmad Malik and
                  He Jiang and
                  Ahmad Ali and
                  Jie Yang and
                  Wei Liu},
  title        = {3D-MMFN: Multi-level multimodal fusion network for 3D industrial image
                  anomaly detection},
  journal      = {Adv. Eng. Informatics},
  volume       = {65},
  pages        = {103284},
  year         = {2025}
}

@inproceedings{0020PZYWW23,
  author       = {Yue Wang and
                  Jinlong Peng and
                  Jiangning Zhang and
                  Ran Yi and
                  Yabiao Wang and
                  Chengjie Wang},
  title        = {Multimodal Industrial Anomaly Detection via Hybrid Fusion},
  booktitle    = {{IEEE/CVF} Conference on Computer Vision and Pattern Recognition, Vancouver, BC, Canada},
  pages        = {8032--8041},
  publisher    = {{IEEE}},
  year         = {2023}
}

@article{chen2024multi,
  title={Multi-view graph contrastive learning for social recommendation},
  author={Chen, Rui and Chen, Jialu and Gan, Xianghua},
  journal={Scientific reports},
  volume={14},
  number={1},
  pages={22643},
  year={2024},
  publisher={Nature Publishing Group UK London}
}

@inproceedings{HuangZLLHZ18,
  author       = {Feiran Huang and
                  Xiaoming Zhang and
                  Chaozhuo Li and
                  Zhoujun Li and
                  Yueying He and
                  Zhonghua Zhao},
  title        = {Multimodal Network Embedding via Attention based Multi-view Variational
                  Autoencoder},
  booktitle    = {Proceedings of the 2018 {ACM} on International Conference on Multimedia
                  Retrieval, Yokohama, Japan},
  pages        = {108--116},
  publisher    = {{ACM}},
  year         = {2018}
}

@article{li2025curegraph,
  title={CureGraph: Contrastive multi-modal graph representation learning for urban living circle health profiling and prediction},
  author={Li, Jinlin and Zhou, Xiao},
  journal={Artificial Intelligence},
  volume={340},
  pages={104278},
  year={2025},
  publisher={Elsevier}
}

@article{holm2025amvae,
  title={amVAE: Age-aware multimorbidity clustering using variational autoencoders},
  author={Holm, Nikolaj Normann and Le, Thao Minh and Fr{\o}lich, Anne and Andersen, Ove and Juul-Larsen, Helle Gybel and Stockmarr, Anders and Venkatesh, Svetha},
  journal={Computers in Biology and Medicine},
  volume={186},
  pages={109632},
  year={2025},
  publisher={Elsevier}
}

@article{cui2025towards,
  title={Towards multimodal foundation models in molecular cell biology},
  author={Cui, Haotian and Tejada-Lapuerta, Alejandro and Brbi{\'c}, Maria and Saez-Rodriguez, Julio and Cristea, Simona and Goodarzi, Hani and Lotfollahi, Mohammad and Theis, Fabian J and Wang, Bo},
  journal={Nature},
  volume={640},
  number={8059},
  pages={623--633},
  year={2025},
  publisher={Nature Publishing Group UK London}
}

@article{rao2025multimodal,
  title={Multimodal generative AI for medical image interpretation},
  author={Rao, Vishwanatha M and Hla, Michael and Moor, Michael and Adithan, Subathra and Kwak, Stephen and Topol, Eric J and Rajpurkar, Pranav},
  journal={Nature},
  volume={639},
  number={8056},
  pages={888--896},
  year={2025},
  publisher={Nature Publishing Group UK London}
}

@inproceedings{gao2015multi,
  title={Multi-view subspace clustering},
  author={Gao, Hongchang and Nie, Feiping and Li, Xuelong and Huang, Heng},
  booktitle={Proceedings of the IEEE international conference on computer vision},
  pages={4238--4246},
  year={2015}
}

@inproceedings{cao2015diversity,
  title={Diversity-induced multi-view subspace clustering},
  author={Cao, Xiaochun and Zhang, Changqing and Fu, Huazhu and Liu, Si and Zhang, Hua},
  booktitle={Proceedings of the IEEE conference on computer vision and pattern recognition},
  pages={586--594},
  year={2015}
}

@inproceedings{kang2020large,
  title={Large-scale multi-view subspace clustering in linear time},
  author={Kang, Zhao and Zhou, Wangtao and Zhao, Zhitong and Shao, Junming and Han, Meng and Xu, Zenglin},
  booktitle={Proceedings of the AAAI conference on artificial intelligence},
  volume={34},
  number={04},
  pages={4412--4419},
  year={2020}
}

@inproceedings{liu2013multi,
  title={Multi-view clustering via joint nonnegative matrix factorization},
  author={Liu, Jialu and Wang, Chi and Gao, Jing and Han, Jiawei},
  booktitle={Proceedings of the 2013 SIAM international conference on data mining},
  pages={252--260},
  year={2013},
}

@inproceedings{zhao2017multi,
  title={Multi-view clustering via deep matrix factorization},
  author={Zhao, Handong and Ding, Zhengming and Fu, Yun},
  booktitle={Proceedings of the AAAI conference on artificial intelligence},
  volume={31},
  number={1},
  year={2017}
}

@article{huang2020auto,
  title={Auto-weighted multi-view clustering via deep matrix decomposition},
  author={Huang, Shudong and Kang, Zhao and Xu, Zenglin},
  journal={Pattern Recognition},
  volume={97},
  pages={107015},
  year={2020},
  publisher={Elsevier}
}

@inproceedings{tang2020cgd,
  title={CGD: Multi-view clustering via cross-view graph diffusion},
  author={Tang, Chang and Liu, Xinwang and Zhu, Xinzhong and Zhu, En and Luo, Zhigang and Wang, Lizhe and Gao, Wen},
  booktitle={Proceedings of the AAAI conference on artificial intelligence},
  volume={34},
  number={04},
  pages={5924--5931},
  year={2020}
}

@article{wang2019gmc,
  title={GMC: Graph-based multi-view clustering},
  author={Wang, Hao and Yang, Yan and Liu, Bing},
  journal={IEEE Transactions on Knowledge and Data Engineering},
  volume={32},
  number={6},
  pages={1116--1129},
  year={2019},
  publisher={IEEE}
}

@inproceedings{JinWDLZ23,
  author       = {Jiaqi Jin and
                  Siwei Wang and
                  Zhibin Dong and
                  Xinwang Liu and
                  En Zhu},
  title        = {Deep Incomplete Multi-View Clustering with Cross-View Partial Sample
                  and Prototype Alignment},
  booktitle    = {{IEEE/CVF} Conference on Computer Vision and Pattern Recognition,
                 Vancouver, BC, Canada},
  pages        = {11600--11609},
  publisher    = {{IEEE}},
  year         = {2023}
}

@inproceedings{PuCC0PHY024,
  author       = {Jingyu Pu and
                  Chenhang Cui and
                  Xinyue Chen and
                  Yazhou Ren and
                  Xiaorong Pu and
                  Zhifeng Hao and
                  Philip S. Yu and
                  Lifang He},
  title        = {Adaptive Feature Imputation with Latent Graph for Deep Incomplete
                  Multi-View Clustering},
  booktitle    = {Thirty-Eighth Conference on Artificial Intelligence, February 20-27, 2024, Vancouver,
                  Canada},
  pages        = {14633--14641},
  publisher    = {{AAAI} Press},
  year         = {2024}
}

@inproceedings{ChaoJC24,
  author       = {Guoqing Chao and
                  Yi Jiang and
                  Dianhui Chu},
  title        = {Incomplete Contrastive Multi-View Clustering with High-Confidence
                  Guiding},
  booktitle    = {Thirty-Eighth Conference on Artificial Intelligence, Vancouver, Canada},
  pages        = {11221--11229},
  publisher    = {{AAAI} Press},
  year         = {2024}
}

@inproceedings{0001DWC0F024,
  author       = {Jie Wen and
                  Shijie Deng and
                  Waikeung Wong and
                  Guoqing Chao and
                  Chao Huang and
                  Lunke Fei and
                  Yong Xu},
  title        = {Diffusion-based Missing-view Generation With the Application on Incomplete Multi-view Clustering},
  booktitle    = {Forty-first International Conference on Machine Learning,
                  Vienna, Austria},
  publisher    = {OpenReview.net},
  year         = {2024}
}

@inproceedings{FengSWGTD24,
  author       = {Wei Feng and
                  Guoshuai Sheng and
                  Qianqian Wang and
                  Quanxue Gao and
                  Zhiqiang Tao and
                  Bo Dong},
  title        = {Partial Multi-View Clustering via Self-Supervised Network},
  booktitle    = {Thirty-Eighth Conference on Artificial Intelligence, Vancouver,
                  Canada},
  pages        = {11988--11995},
  publisher    = {{AAAI} Press},
  year         = {2024}
}

@inproceedings{XuL0PMS022,
  author       = {Jie Xu and
                  Chao Li and
                  Yazhou Ren and
                  Liang Peng and
                  Yujie Mo and
                  Xiaoshuang Shi and
                  Xiaofeng Zhu},
  title        = {Deep Incomplete Multi-View Clustering via Mining Cluster Complementarity},
  booktitle    = {Thirty-Sixth Conference on Artificial Intelligence},
  pages        = {8761--8769},
  publisher    = {{AAAI} Press},
  year         = {2022}
}

@inproceedings{Xu0W0Z0024,
  author       = {Jie Xu and
                  Yazhou Ren and
                  Xiaolong Wang and
                  Lei Feng and
                  Zheng Zhang and
                  Gang Niu and
                  Xiaofeng Zhu},
  title        = {Investigating and Mitigating the Side Effects of Noisy Views for Self-Supervised
                  Clustering Algorithms in Practical Multi-View Scenarios},
  booktitle    = {{IEEE/CVF} Conference on Computer Vision and Pattern Recognition,
                 Seattle, WA, USA},
  pages        = {22957--22966},
  publisher    = {{IEEE}},
  year         = {2024}
}

@article{yang2025automatically,
  title={Automatically Identify and Rectify: Robust Deep Contrastive Multi-view Clustering in Noisy Scenarios},
  author={Yang, Xihong and Wang, Siwei and Wang, Fangdi and Jin, Jiaqi and Liu, Suyuan and Liu, Yue and Zhu, En and Liu, Xinwang and Jin, Yueming},
  journal={arXiv preprint arXiv:2505.21387},
  year={2025}
}

@inproceedings{SunLRDP025,
  author       = {Yuan Sun and
                  Yongxiang Li and
                  Zhenwen Ren and
                  Guiduo Duan and
                  Dezhong Peng and
                  Peng Hu},
  title        = {{ROLL:} Robust Noisy Pseudo-label Learning for Multi-View Clustering with Noisy Correspondence},
  booktitle    = {{IEEE/CVF} Conference on Computer Vision and Pattern Recognition,
                  Nashville, TN, USA},
  pages        = {30732--30741},
  publisher    = {Computer Vision Foundation / {IEEE}},
  year         = {2025}
}

@inproceedings{zhang2025incomplete,
  title={Incomplete Multi-view Clustering via Diffusion Contrastive Generation},
  author={Zhang, Yuanyang and Lin, Yijie and Yan, Weiqing and Yao, Li and Wan, Xinhang and Li, Guangyuan and Zhang, Chao and Ke, Guanzhou and Xu, Jie},
  booktitle={Proceedings of the AAAI Conference on Artificial Intelligence},
  volume={39},
  number={21},
  pages={22650--22658},
  year={2025}
}

@article{zhang2025multi,
  title={Multi-branch Space Sharing Feature Aggregation for contrastive multi-view clustering},
  author={Zhang, Yuanyang and Yan, Weiqing and Tang, Chang and Zhou, Wujie and Jin, Jian},
  journal={Pattern Recognition},
  pages={111704},
  year={2025},
  publisher={Elsevier}
}

\end{document}